\documentclass[10pt,twocolumn,letterpaper]{article}

\usepackage{cvpr}
\usepackage{times}
\usepackage{epsfig}
\usepackage{graphicx}
\usepackage{amsmath}
\usepackage{amssymb}

\usepackage{multirow}
\usepackage[pagebackref=true,breaklinks=true,letterpaper=true,colorlinks,bookmarks=false]{hyperref}

 \cvprfinalcopy % *** Uncomment this line for the final submission

\def\cvprPaperID{840} % *** Enter the CVPR Paper ID here

\ifcvprfinal\fi
\begin{document}

%%%%%%%%% TITLE
\title{Deep Contrast Learning for Salient Object Detection}
\author{Guanbin Li \hspace{1.0in} Yizhou Yu\\
Department of Computer Science, The University of Hong Kong\\
{\tt\small \{gbli, yzyu\}@cs.hku.hk}%https://sites.google.com/site/ligb86/mdfsaliency/ }
%\and
}

\maketitle
%\thispagestyle{empty}

%%%%%%%%% ABSTRACT
\begin{abstract}
   Salient object detection has recently witnessed substantial progress due to powerful features extracted using deep convolutional neural networks (CNNs). However, existing CNN-based methods operate at the patch level instead of the pixel level. Resulting saliency maps are typically blurry, especially near the boundary of salient objects. Furthermore, image patches are treated as independent samples even when they are overlapping, giving rise to significant redundancy in computation and storage.
   %Currently the leading salient object detection approaches focus on learning saliency models or extracting CNN features in the unit of an image region. This strategy introduces artificial boundaries on the saliency map. Besides, image regions are processed in an independent way without sharing computation and it requires to compute thousands of networks on a single image, making both training and testing very inefficient.
   %that capture the semantic context in the image. These approaches generally treat an image region~(i.e. object proposal, superpixel) as a basic unit in deep feature extraction or learning their saliency models. Regions are processed in an independent way without sharing computation.
   In this paper, we propose an end-to-end deep contrast network to overcome the aforementioned limitations. Our deep network consists of two complementary components, a pixel-level fully convolutional stream and a segment-wise spatial pooling stream. The first stream directly produces a saliency map with pixel-level accuracy from an input image. The second stream extracts segment-wise features very efficiently, and better models saliency discontinuities along object boundaries. Finally, a fully connected CRF model can be optionally incorporated to improve spatial coherence and contour localization in the fused result from these two streams. Experimental results demonstrate that our deep model significantly improves the state of the art.
   %In this paper, we formulate salient object detection as a regression problem using end to end deep learning techniques. A novel deep contrast neural network is proposed to inference the saliency value from high-level contrast information of both pixel and regional level. We then propose a refinement method based on fully connected CRF to enhance the spatial coherence of our saliency results. We extensively evaluate our proposed method on five public datasets, and experimental results show significant and consistent improvements and competitive runtime speed over the state-of-the-arts methods.

   %On pixelwise contrast learning, a multi-scale fully convolutional neural network~(MS-FCN) is applied to regress end to end from an input image to an initial saliency map. A convolutional feature masking layer is added on top of last convolutional layer of MS-FCN for segment feature extraction. On regional contrast learning, two fully connected layers are designed to map each superpixel with concatenated three scales segment features to a saliency value, ending up with regional saliency map. We further propose a multi-scale fully convolutional neural network to extract the saliency contour of

\end{abstract}

%%%%%%%%% BODY TEXT
\section{Introduction}
Visual saliency aims at identifying the most visually distinctive parts in an image, and has received increasing interest in recent years. Though early work primarily focused on predicting eye-fixations in images, research has shown that salient object detection, which emphasizes object-level integrity of saliency prediction results, is more useful and can serve as a pre-processing step for a variety of computer vision and image processing tasks including content-aware image editing~\cite{avidan2007seam}, object detection~\cite{navalpakkam2006integrated}, image classification~\cite{WYW13}, %visual tracking~\cite{mahadevan2009saliency},
person re-identification~\cite{bi2014person} and video summarization~\cite{ma2002user}. Despite recent progress, salient object detection remains a challenging problem that calls for more accurate solutions.

Results from perceptual research~\cite{einhauser2003does,parkhurst2002modeling} indicate that visual contrast is the most important factor in visual saliency. Various conventional saliency detection algorithms based on local or global contrast cues~\cite{cheng2015global,yang2013saliency} have been successfully developed. In previous work, visual contrast is exemplified by contrast in various types of handcrafted low-level features (e.g., color, intensity and texture) at the pixel or segment level. Though handcrafted features tend to perform well in standard scenarios, they are not sufficiently robust for all challenging cases. For example, local contrast features may fail to detect homogenous regions inside salient objects while global contrast suffers from complex background.
Although machine learning based saliency models have been developed \cite{lu2014learning,jiang2013salient,liu2011learning,mai2013saliency}, they are primarily for integrating different handcrafted features~\cite{jiang2013salient} or fusing multiple saliency maps generated from different methods~\cite{mai2013saliency}.

To obtain more robust features than handcrafted ones for salient object detection, deep convolutional neural networks (CNNs) have recently been employed, achieving substantially better results than previous state of the art~\cite{LiYu15,zhao2015saliency,wang2015deep}. In addition to improved robustness, features extracted using CNNs contain more high-level semantic information since those CNNs were typically pre-trained on datasets for visual recognition tasks. However, in all these methods, CNNs are all operated at the patch level instead of the pixel level, and each pixel is simply assigned the saliency value of its enclosing patch. As a result, saliency maps are typically blurry without fine details, especially near the boundary of salient objects. Furthermore, all image patches are treated as independent data samples for classification or regression even when they are overlapping. As a result, these methods usually have to run a CNN at least thousands of times (once for every patch) to obtain a complete saliency map. This gives rise to significant redundancy in computation and storage, and makes both training and testing very space and time consuming. For example, training a patch-oriented CNN model for saliency detection takes over 2 GPU days and requires hundreds of gigabytes of storage for the 5000 images in the MSRA-B dataset.
%For testing, it take a magnitude of ten seconds to compute thousands times of networks on a single image to generate a dense saliency map.
%In this paper, we propose an end to end deep contrast network that fixes all these disadvantages of the current CNN based saliency detection methods.

In this paper, inspired by a recent trend of developing fully convolutional neural networks %end-to-end deep models
for pixel labeling problems~\cite{long2014fully,chen2014semantic,xie2015holistically}, we propose an end-to-end deep contrast network to overcome the aforementioned limitations of recent CNN-based saliency detection methods. \color{black} Here, ``end-to-end'' means that our deep network only needs to be run on the input image once to produce a complete saliency map with the same pixel resolution as the input image. \color{black} Our deep network consists of a pixel-level fully convolutional stream and a segment-level spatial pooling stream. In the fully convolutional stream, we design a multi-scale fully convolutional network (MS-FCN), which takes the raw image as input and directly produces a saliency map with pixel-level accuracy. Our MS-FCN can not only generate effective semantic features across different scales, but also capture subtle visual contrast among multi-scale feature maps for saliency inference. The segment-level spatial pooling stream generates another saliency map at the superpixel level by performing spatial pooling and saliency estimation over superpixels. This stream extracts segment-wise features very efficiently from MS-FCN by masking an intermediate feature map computed for the entire image. The saliency maps from both streams are fused at the end. %\color{red}(delete) Our MS-FCN can also be re-trained to generate a contour map for salient objects. This contour map can be used to improve contour localization in the fused saliency map via a fully connected CRF. \color{black}

In summary, this paper has the following contributions:
\begin{itemize}
\item We introduce an end-to-end deep contrast network for salient object detection. It consists of a fully convolutional stream and a segment-wise spatial pooling stream. A training scheme is designed to learn the weights in both streams of this deep network. The fused saliency map from these two streams is further refined with a fully connected CRF for better spatial coherence and contour localization.
\item We propose a multi-scale fully convolutional network as the first stream in our deep contrast network to infer a pixel-level saliency map directly from the raw input image. \color{black} This model can not only infer semantic properties of salient objects, but also capture visual contrast among multi-scale feature maps. \color{black}
    %This fully convolutional network can also be re-trained to infer a contour map for salient objects.
\item We also design a segment-wise spatial pooling stream as the second stream in our framework. This stream efficiently extracts segment-wise features, and accurately models visual contrast between regions and saliency discontinuities along region boundaries.
    %feature masking (SFM) layer to instantaneous masking a feature vector for each segment from the last convolutional feature map of our proposed MS-FCN. Then another saliency map is inference by learning the regional contrast information based on the incorporated multiscale segment features. This saliency map is iterately used to update our MS-FCN.
%\item The proposed MS-FCN framework is modified to generate a saliency edge map for each testing image, which is further used as a post-processing in a fully connected CRF framework to generate a final boundary-preserving saliency map with spatial coherence.
\end{itemize}

%Similar to the R-CNN method for object detection [8], these
%methods need to run the forward pass of a deep neural network hundreds or thousands times to inference an entire saliency map. This is very time-consuming even on computers with top-notch GPUs.
%-------------------------------------------------------------------------
\section{Related Work}
%\subsection{Salient Object Detection}
Salient object detection can be performed either in a bottom-up fashion using low-level features ~\cite{gao2007bottom,achanta2009frequency,liu2011learning,klein2011center,perazzi2012saliency,yang2013saliency,jiang2013salient,zhu2014saliency,cheng2015global} or in a top-down fashion via the incorporation of high-level knowledge~\cite{judd2009learning,chang2011fusing,goferman2012context,shen2012unified,liu2014adaptive,jia2013category,li2014secrets}.
Since this paper is focused on visual saliency based on deep learning, we discuss relevant work in this context below.

Recently, machine learning and artificial intelligence have been revolutionized by deep convolutional neural networks, which have set new state of the art on a number of visual recognition tasks, including image classification~\cite{krizhevsky2012imagenet}, object detection~\cite{girshick2014rich}, scene classification~\cite{yan2015hd} and scene parsing~\cite{farabet2013learning}, closing the gap to human-level performance.
%Razavian {\em et al.}~\cite{razavian2014cnn} and Donahue {\em et al.}~\cite{donahue2013decaf} pointed out that off-the-shelf features learned by CNN trained on the ImageNet dataset~\cite{deng2009imagenet} can be effectively adapted to generic tasks.
There have also been attempts to apply deep learning to salient object detection. Li {\em et al.}~\cite{LiYu15} trained a deep neural network for deriving a saliency map from multiscale features extracted using deep convolutional neural networks. Wang {\em et al.}~\cite{wang2015deep} adopted a deep neural network (DNN-L) to learn local patch features for each centered pixel. In~\cite{zhao2015saliency}, both global context and local context are utilized and integrated into a deep learning based pipeline for saliency detection. However, all these methods treat local image patches as independent training and testing samples.
%Each patch is either padded or enlarged to the size of a complete image to meet the requirement of a traditional deep convolutional neural network for image classification.
Since sharing computation among overlapping patches is not considered, there is a great deal of redundancy in feature computation, which gives rise to high computational cost for both training and testing.
%Inspired by a recent trend of developing end-to-end deep models for pixel labeling problems~\cite{long2014fully,chen2014semantic,xie2015holistically}, we propose to develop an end-to-end model for salient object detection in this paper.
This limitation can be potentially overcome by recent end-to-end deep networks, which have been proven a success in semantic segmentation~\cite{long2014fully,chen2014semantic}. However, directly applying existing %end-to-end
fully convolutional network architecture to salient object detection would not be most appropriate because a standard fully convolutional % end-to-end 
model is not particularly good at capturing subtle visual contrast in an image.
%is capable of grasping the semantic information at the pixel level using the high-level knowledge learned during training, but it does not have the knowledge to tell which part of an image is salient. Visual contrast modeling is actually the essence of salient object detection.
Therefore, our paper focuses on discovering high-level visual contrast in an end-to-end mode, and experimental results demonstrate that our proposed deep model can significantly improve the current state of the art. This paper can be viewed as the first piece of work that aims to discover visual contrast information inside an image using end-to-end convolutional neural networks.

%Noted in a recent yet unpublished work, the author directly learn the saliency from a semantic view which only make use of the semantic information but not dip into the high-level contrast learning. Our framework mainly focus on discovering the high-level contrast relationship in an end to end mode, and experimental results demostrates that our proposed method can substantially outperform outperforms the current states of the art. Our work can be viewed as the first work to explore the contrast information inside an image using end to end convolutional neural network.

\begin{figure}[ht]
\begin{center}
   \includegraphics[width=\columnwidth]{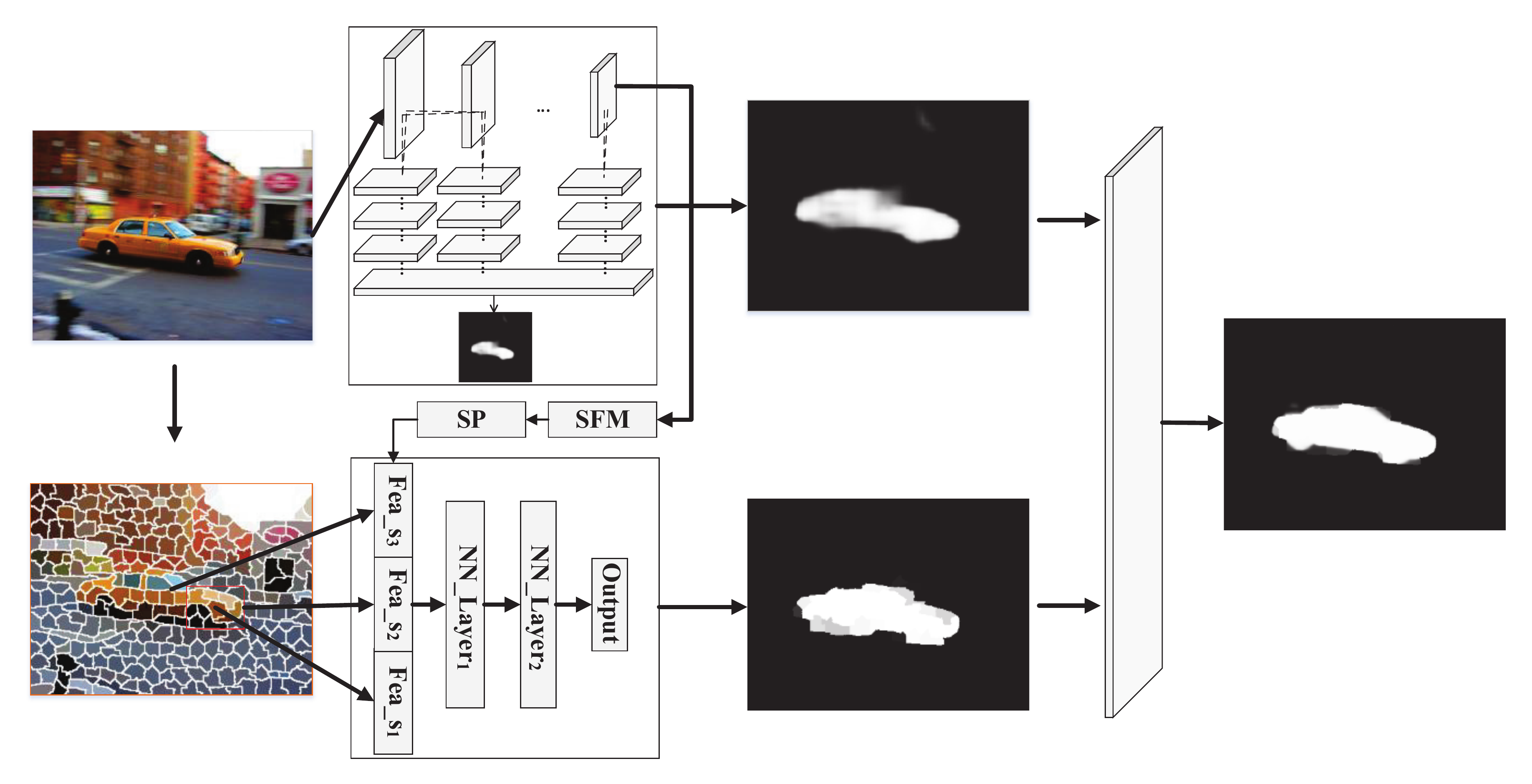}
\end{center}
   \caption{Two streams of our deep contrast network.}
\label{fig:dcn}
\end{figure}

\begin{figure}[ht]
\begin{center}
   \includegraphics[width=\columnwidth]{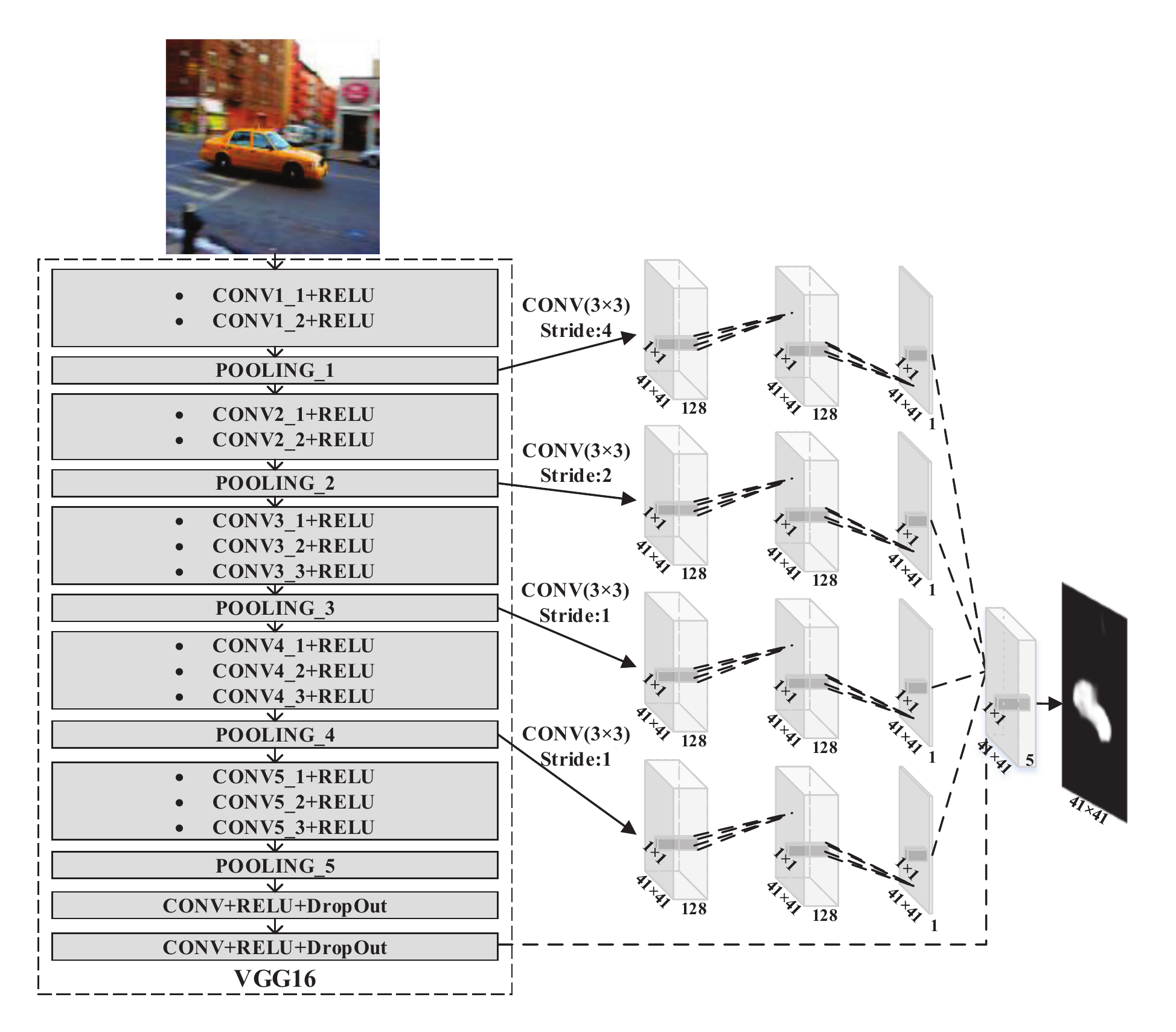}
\end{center}
   \caption{The architecture of multi-scale fully convolutional network.}
\label{fig:ms-fcn}
\end{figure}

\section{Deep Contrast Network}\label{sec:deep_contrast_network}
As shown in Fig.~\ref{fig:dcn}, the architecture of our deep contrast network for salient object detection consists of two complementary components, a fully convolutional stream and a segment-wise spatial pooling stream. The fully convolutional stream is a multi-scale fully convolutional network~(MS-FCN), which generates a saliency map $S_1$ with one eighth resolution of the raw input image by exploiting visual contrast across multiscale convolutional layers. The segment-wise spatial pooling stream generates a saliency map at the superpixel level by performing spatial pooling and saliency estimation over individual superpixels. The saliency maps from both streams are fused at the end \color{black} through an extra convolutional layer with $1\times 1$ kernels in our deep network \color{black} to produce the final saliency map. The weights in this fusion layer are learned during training. %MS-FCN is a modification of VGGNet\cite{simonyan2014very} as described in section~\ref{sec:ms-fcn}. The segment stream which is designed for regional contrast learning starts with a segment feature masking layer~(SFM) on top of the last convolutional layer~(Conv5\_3) of MS-FCN. We first obtain a superpixel segmentation on the raw image. CFM together with a SPP~\cite{he2014spatial} layer act as a role of generating a fixed-length feature vector for an arbitrary sized region. For each superpixel, we extract three feature vectors from three nested and increasingly larger regions as described in~\cite{LiYu15} , which are respectively the considered superpixel, the union region of its immediate neighboring superpixels, and the entire image. This three feature vectors are concatenated and fed into two fully connected layers. The cross-entropy loss layer is added to infer the probability of a region being salient and the saliency map $S_2$ is thus generated by transferring all segments saliency value to all corresponding pixels. Finally, $S_1$ and $S_2$ are stacked and fed to a weight fusion layer~(1*1 convolution layer) to generate a fused saliency map $S$.

\subsection{Multi-Scale Fully Convolutional Network}\label{sec:ms-fcn}
In the fully convolutional stream, we aim to design an end-to-end convolutional network that can be viewed as a regression network mapping an input image to a pixel-level saliency map. To conceive such an end-to-end architecture, we have the following considerations. First, the network should be deep enough to produce multi-level features for detecting salient objects at different scales. Second, the network should be able to discover subtle visual contrast across multiple maps holding deep features at different scales. Last but not the least, fine-tuning an existing deep model is much desired since we do not have enough training images to train such a deep network from scratch.

We chose VGG16~\cite{simonyan2014very} as our pre-trained network and modified it to meet our requirements. To re-purpose it into a dense image saliency prediction network, the two fully connected layers of VGG16 are first converted into convolutional ones with $1\times 1$ kernel as described in \cite{long2014fully}. However directly evaluating the resulting network in a convolutional manner yields a very sparse prediction map with a 32-pixel stride since the original VGG16 network has 5 pooling layers each of which has stride 2. To make the prediction map denser, we skip subsampling in the last two max-pooling layers to maintain an 8-pixel stride after the last pooling layer. To retain the original receptive field in the convolutional layers that follow, we use the ``hole algorithm" to introduce zeros to increase the size of their convolutional kernels. The ``hole algorithm", which is also called \`a trous algorithm, was originally developed for efficient computation of the undecimated wavelet transform~\cite{mallat1999wavelet}, and has recently been implemented in Caffe~\cite{chen2014semantic,li2014highly} to efficiently compute dense CNN feature maps at any target subsampling rate without introducing any approximation.
%Given an 1-D input signal $x[i]$ with a filter $w[i]$ of length $K$, the output signal $y[i]$ of convolution with hole algorithm is calculated as
%\begin{equation}
%y[i]=\sum_{k=1}^{K}x[i+s\cdot k]\cdot w[k],
%\end{equation}
%where $s$ is the stride.
This hole algorithm helps us keep the kernels intact, and a convolution now sparsely samples the input feature map using a stride of 2 or 4 pixels (2-pixel stride in the three convolutional layers after the penultimate pooling layer and 4-pixel stride in the last two converted $1\times 1$ convolutional layers after the final pooling layer). For our experiments, we followed the implementation of the published DeepLab code~\cite{chen2014semantic} %, which is implemented based on the publicly available Caffe platform~\cite{jia2014caffe}
and added the option to sparsely sample the underlying feature map to the `im2col' function. % in the implementation of the published DeepLab code~\cite{chen2014semantic}, which is implemented based on the publicly available Caffe platform.
`im2col' is a function implemented in Caffe to convert multi-channel feature maps to vectorized patches for improving the efficiency of convolutions.

VGG16 has five pooling and downsampling layers, each of which has an increasingly larger receptive field containing contextual information. To design a deep network that is capable of discovering visual contrast crucial in saliency inference, we further develop a multiscale version of the above %end-to-end
fully convolutional extension of VGG16. As shown in Fig.~\ref{fig:ms-fcn}, we connect three extra convolutional layers to each of the first four max-pooling layers of VGG16. The first extra layer has $3\times 3$ kernels and 128 channels, the second extra layer has $1\times 1$ kernels and 128 channels, and the third extra layer (output feature map) has a $1\times 1$ kernel and a single channel. To make the output feature maps of the four sets of extra convolutional layers have the same size (8$\times$ subsampled resolution), the stride of the first layer in these four sets are set to 4, 2, 1, and 1, respectively. Although the resulting four output maps have the same size, they are generated using receptive fields with different sizes and hence represent contextual features at 4 different scales. We further stack these four feature maps together with the final output map of the above end-to-end extension. The stacked feature maps (5 channels) are fed into a final convolutional layer with a $1\times 1$ kernel and a single output channel, which is the inferred saliency map. The sigmoid activation function is used in the final layer. Although the output saliency map is of $8\times$ subsampled resolution, they are smooth enough and allow us to use simple bilinear interpolation to make their resolution the same as that of the original input image at a negligible computational cost.
%In order to concatenate the feature map with the corresponding one generated from our segment stream for a fusion weight learning, we simply resize the output saliency map to the size of the orginal image using bilinear interpolation, and
We call this resized saliency map $S_1$.

Note that the method in \cite{long2014fully} does not use the ``hole algorithm" and produces very coarse maps (subsampled by a factor of 32), which motivate the use of trained deconvolution layers. The incorporation of deconvolution layers significantly increases the complexity and training time of their network. Experimental results also show that convolution with the ``hole algorithm" can generate better results than trained deconvolution layers~\cite{chen2014semantic}.

\subsection{Segment-Level Saliency Inference}
%Though multi-Scale fully convolutional network can capture some contrast feature from multi-scale feature maps. However, this is not enough since all activations have the same receptive size on the same feature map and both convolution and pooling are operated on a regular mode.
Salient objects often have irregular shapes and the corresponding saliency map has discontinuities along object boundaries. Our multiscale fully convolutional network operates at a subsampled pixel level without explicitly modeling such saliency discontinuities. To better model visual contrast between regions and visual saliency along region boundaries, we design a segment-wise spatial pooling stream in our network.

We first decompose the raw input image into a set of superpixels, and call each superpixel a segment. A mask is computed for every segment in the feature map generated from the last true convolutional layer (Conv5\_3) of MS-FCN as follows. Since each activation in Conv5\_3 is controlled by a receptive field in the input image, we first project every activation to the center of its receptive field as in~\cite{girshickICCV15fastrcnn,dai2015convolutional}. For each segment in the input image, we first generate a binary mask with the same size as its bounding box. In this mask, pixels inside the segment are labeled `$1$' while others are labeled `$0$'. Each label in the binary mask is first assigned to the nearest center of receptive field and then backprojected onto Conv5\_3. Thus, each activation in Conv5\_3 collects multiple binary labels backprojected from its receptive field. The collected binary labels at each activation are first averaged and then thresholded by 0.5, yielding a corresponding binary segment mask on Conv5\_3, where pixels within the segment can be easily identified according to this mask.
Note that feature maps generated from Conv5\_3 have 8-pixel strides in our MS-FCN instead of 32-pixel ones in the original VGG16 network since subsampling was skipped in the last two max-pooling layers as described in Section~\ref{sec:ms-fcn}. Therefore, the resolution of the feature map generated from Conv5\_3 is sufficient for segment masking.

Since segments on Conv5\_3 have variable size, to produce a fixed-length feature vector, we further perform spatial pooling (SP) over a fixed grid as with ~\cite{he2014spatial}. We divide the bounding box of a segment on Conv5\_3 into $h\times w$ cells. Let the size of the bounding box be $H\times W$. Spatial pooling is performed within each cell with $H/h \times W/w$ pixels. Afterwards, the aggregated feature vector of each segment has $h\times w\times C$ dimensions, where $C$ is the number of channels of the feature map generated by Conv5\_3.

To discover segment-level visual contrast, for each segment, we obtain three spatially aggregated feature vectors from three nested and increasingly larger windows, which are respectively the bounding box of the considered segment, the bounding box of its immediate neighboring segments, and the entire map from Conv5\_3 (with the considered segment masked out to indicate the position of the segment in the map). Finally, the three aggregated feature vectors are concatenated and fed into two fully connected layers. The output of the second fully connected layer is fed into the output layer, which uses the sigmoid function to perform logistic regression to produce a distribution over binary saliency labels. We call the saliency map generated in this way $S_2$.

This segment-wise spatial pooling stream of our network is in fact an accelerated version of the method in \cite{LiYu15}. Although they share similar strategies for multiscale feature extraction, our method is much more efficient because convolutional feature maps only need to be computed once for the entire image and afterwards, local features for thousands of segments from the same image can be masked out instantaneously. Moreover, our model also achieves better results as segment features are extracted from our multiscale fully convolutional network, which has been fine-tuned for salient object detection, instead of from the original VGG16 model for image classification.
%Recently, end-to-end deep learning system has been successfully used in the area of semantic segmentation by either transforming a standard CNN into a fully convolutional network~\cite{long2014fully,chen2014semantic} or learning a multi-layer deconvolution network\cite{noh2015learning}. Difference from semantic segmentation,

\subsection{Deep Contrast Network Training}\label{sec:dcn_train}
Given training images and their superpixels, we first train the neural network in the second stream alone to obtain its initial weights. Segment features are extracted using the original VGG16 network pre-trained over the ImageNet dataset~\cite{deng2009imagenet}. After this initialization, we fine-tune the two streams of our deep contrast network in an alternating manner. We first fix the parameters in the second stream and train the first stream for one epoch. During this process, the weights for fusing the saliency maps ($S_1$ and $S_2$) from the two streams as well as the parameters in the multiscale fully convolutional network are updated using stochastic gradient descent.
Then we fix the parameters in the first stream and fine-tune the neural network in the second stream for one epoch using groundtruth saliency maps. Segment features are extracted using the updated VGG16 network embedded in the first stream. We typically alternate the above two steps 8 times (16 epochs in total) before the whole fine-tuning process converges.
%More specifically, given an image, we first generate $S_1$ and $S_2$ independently from the two streams of the network. These two saliency maps are integrated through a fusion layer which performs a weighted average of the two input maps. The weights for fusing these two maps are also trained.

The loss function for fine-tuning the deep contrast network (the first stream) and the fusing weights is the cross entropy between the ground truth and the fused saliency map ($S$):
\begin{equation}
\begin{aligned}
L=&-\beta_i\sum_{i=1}^{|I|}G_i\log P\left ( S_i=1|I_i,W \right ) \\& -  \left ( 1-\beta_i \right )\sum_{i=1}^{|I|}\left ( 1-G_i \right )\log P\left ( S_i=0|I_i,W \right ),
\end{aligned}
\end{equation}
%\begin{equation}
%L= \frac{-1}{\left | G \right |}\sum_{i=1}^{\left | G \right |}G_i\log Pr\left (G_i=1|,I_i,W\right)+\left(1-G_i\right)\log \left(Pr\left(G_i=0|,I_i,W\right)\right),
%\end{equation}
where $G$ is the groundtruth label, $W$ denotes the collection of all network parameters in MS-FCN and the fusion layer, $\beta_i$ is a weight balancing the number of salient pixels and unsalient ones, and $|I|$, $|I|\_$ and $|I|_+$ denote the total number of pixels, unsalient pixels and salient pixels in image $I$, respectively. Then $\beta_i=\frac{|I|\_}{|I|}$ and $1-\beta_i=\frac{|I|_+}{|I|}$.
%During our training process, the gradient of the loss is backpropagated to update the fusing weights as well as the parameters in the fully convolutional network.
When fine-tuning the second stream, its parameters are updated by minimizing the squared prediction errors accumulated over all segments from all training images. %After each parameter update, $S_1$ and $S_2$ for an input image are computed using the updated fully convolutional network.

%As illustrated in Fig.~Z, the saliency map $S_1$ generated from the fully convolutional stream of our network can reliably predict the presence and rough position of salient objects in an image. However, it is typically very coarse and especially not able to pricisely position the salienct object boundaries. $S_2$ generated from the second stream can be a good complementary since it assigns saliency scores a segment and most of the time the edges between each segment can well preserve the original image boundaries, though it lacks pixelwise precise prediction and inevitably bring in some artificial boundaries. The fused saliency map $S$ is therefore a very good comprimise and performs much better than the current state-of-the-arts.

\section{The Complete Algorithm}

\subsection{Superpixel Segmentation}
We aim to decompose the input image into non-overlapping segments.
%In order to better reduce the artificial boundaries on the generated saliency map, each segment should be a perceptually homogeneous region while at the same time, strong contours and edges in the image should be well preserved.
In this paper, we use a slightly modified version of the SLIC algorithm~\cite{achanta2010slic}, which uses geodesic image distance~\cite{criminisi2010geodesic} during K-means clustering in the CIELab color space. As discussed in~\cite{wang2013structure},  geodesic distance based superpixels can guarantee connectivity while well preserve edges in the image. In our experiments, we have found that the final saliency detection performance does not vary much when the number of superpixels is between 200 and 300. And the performance becomes slightly worse when the number of superpixels is fewer than 200 or more than 300.

\begin{figure}[b]
\begin{center}
   \includegraphics[width=\columnwidth]{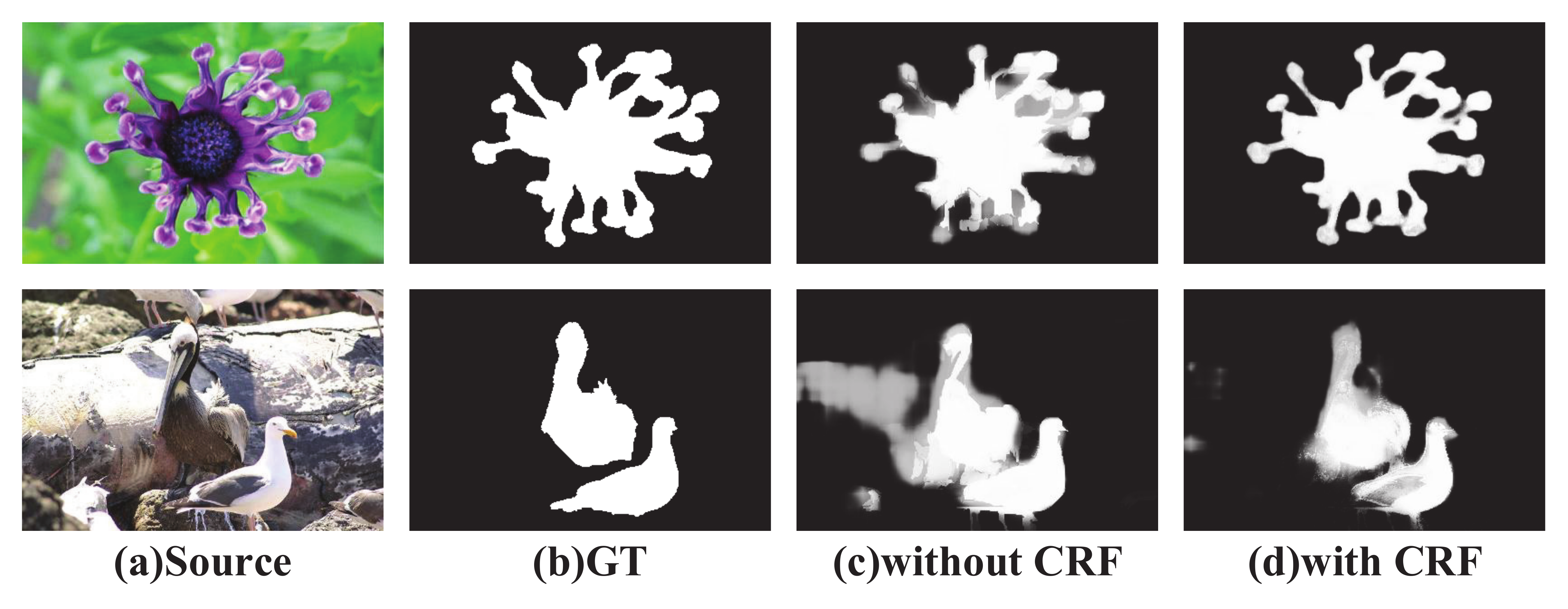}
\end{center}
   \caption{Comparison of saliency detection results with and without CRF.}
\label{fig:crf_effect}
\end{figure}

\subsection{Spatial Coherence}\label{sec:spatial_coherence}
Since both streams in our deep contrast network assign saliency scores to individual pixels or segments without considering the consistency of saliency scores among neighboring pixels and segments, we propose a pixelwise saliency refinement model based on a fully connected CRF~\cite{krahenbuhl2012efficient} to improve spatial coherence. This model solves a binary pixel labeling problem, and employs the following energy function,
\begin{equation}
E\left( L \right) = -\sum_{i}\log P\left(l_i\right)+\sum_{i,j}\theta_{ij}\left(l_i, l_j\right),
\end{equation}
where $L$ represents a binary label (salient or not salient) assignment for all pixels. $P(l_i)$ is the probability of pixel $x_i$ having label $l_i$, which indicates the likelihood of pixel $x_i$ being salient. Initially, $P(1)=S_i$ and $P(0)=1-S_i$, where $S_i$ is the saliency score at pixel $x_i$ from the fused saliency map $S$.
$\theta_{ij}\left(l_i,l_j\right)$ is a pairwise potential and defined as follows,
%\begin{equation}
%\begin{split}
%\theta_{ij}=\mu\left(l_i,l_j\right)\Bigg[ \omega_1\exp\Bigg(-\frac{\left \|p_i-p_j  \right \|^2}{2\sigma_\alpha^2}-\frac{\left \|I_i-I_j \right \|^2}{2\sigma_\beta^2}\\ - \frac{\left \|v_i-v_j \right \|^2}{2\sigma_\gamma ^2}\Bigg) + \omega_2\exp\left(-\frac{\left \|p_i-p_j \right\|^2}{2\sigma_\gamma^2}\right)\Bigg],
%\end{split}
%\end{equation}

\begin{equation}
\begin{split}
\theta_{ij}=\mu\left(l_i,l_j\right)\Bigg[ \omega_1\exp\Bigg(-\frac{\left \|p_i-p_j  \right \|^2}{2\sigma_\alpha^2}-\frac{\left \|I_i-I_j \right \|^2}{2\sigma_\beta^2}\Bigg)  + \\\omega_2\exp\left(-\frac{\left \|p_i-p_j \right\|^2}{2\sigma_\gamma^2}\right)\Bigg],
\end{split}
\end{equation}

where $\mu\left(l_i,l_j\right) = 1$ if $l_i \neq l_j$, and zero otherwise. $\theta_{ij}$ involves two kernels. The first kernel depends on pixel positions ($p$) and pixel intensities ($I$)%and contour-based feature vectors ($v$) generated in Section~\ref{sec:saliency_contour}
. This kernel encourages nearby pixels with similar colors %and without intervening salient object contours
to take similar saliency scores. The degree of influence by color similarity and spatial closeness %and salient object contours
is controlled by three parameters ($\sigma_\alpha$ and $\sigma_\beta$%and $\sigma_\gamma$
), respectively. %The second kernel depends on both pixel position and generated feature vector in section~\ref{sec:saliency_contour}, which encourages pixels without strong saliency boundary crossing to take close saliency value. The degree color similarity and closeness are controlled by parameters~$\theta_\gamma$ and $\theta_\delta$~respectively.
The second kernel aims at removing small isolated regions.

Energy minimization is based on a mean field approximation to the CRF distribution, and high-dimensional filtering can be utilized to speed up the computation. In this paper, we use the publicly available implementation of \cite{krahenbuhl2012efficient} to minimize the above energy, and it takes less than 0.5 second on an image with $300\times 400$ pixels. At the end of energy minimization, we generate a saliency map using the posterior probability of each pixel being salient. We call the generated saliency map $S_{crf}$. As shown in Fig.~\ref{fig:crf_effect}, the saliency maps generated from the proposed method without CRF are fairly coarse and the contours of salient objects may not be well preserved. The proposed saliency refinement model can not only generate smoother results with pixelwise accuracy but also well preserve salient object contours. A quantitative study of the effectiveness of the saliency refinement model can be found in Section~\ref{sec:effectiveness_spatial_coherence}.

\begin{figure*}[ht]
\begin{center}
   \includegraphics[width=\textwidth]{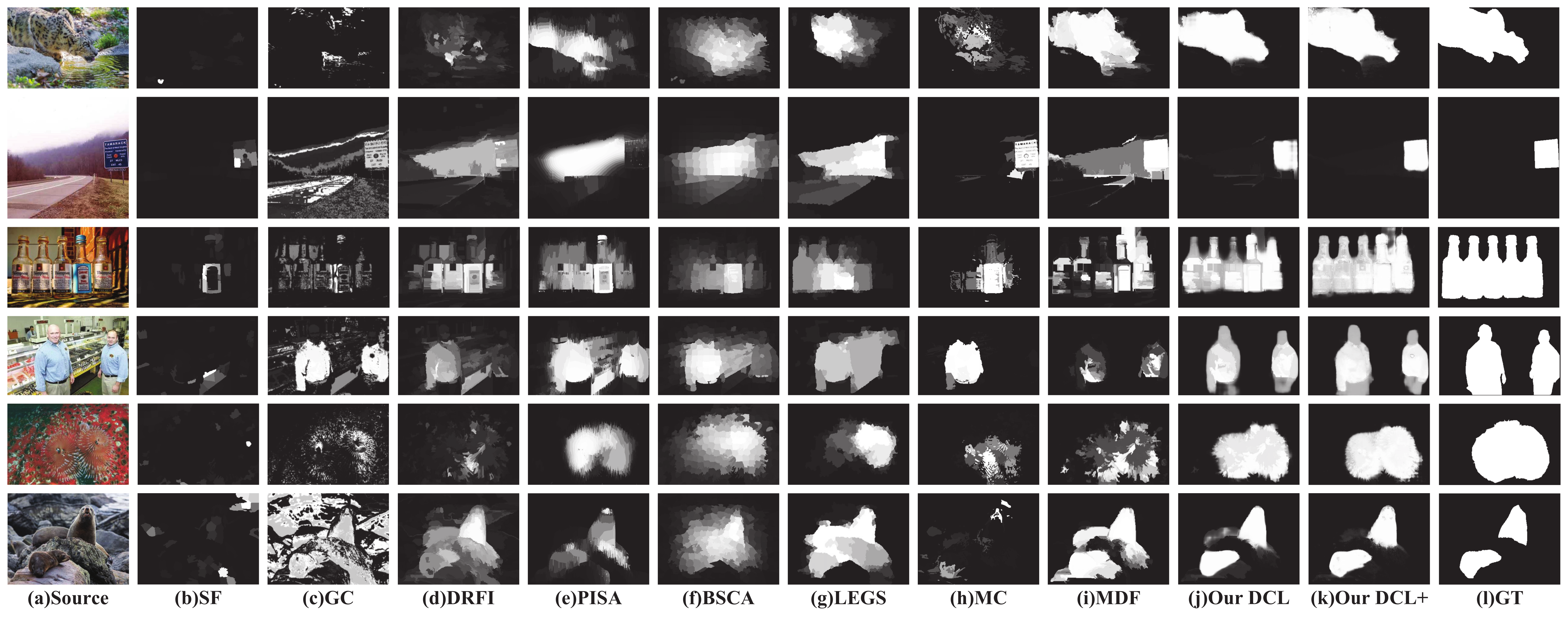}
\end{center}
   \caption{Visual comparison of saliency maps generated from state-of-the-art methods, including our DCL and DCL$^+$. The ground truth (GT) is shown in the last column. DCL$^+$ consistently produces saliency maps closest to the ground truth. %We compare DCL against saliency filters~(SF)~\cite{perazzi2012saliency}, global contrast~(GC)~\cite{cheng2015global}, hierarchical saliency~(HS)~\cite{yan2013hierarchical}, discriminative regional feature integration (DRFI)~\cite{jiang2013salient}, pixelwise image saliency aggregating~(PISA)~\cite{pisa15PAMI}, single-layer cellular automata~(BSCA)~\cite{qin2015saliency},  local estimation and global search based deep networks~(LEGS)~\cite{wang2015deep}, multi-context deep learning~(MC)~\cite{zhao2015saliency}, and multiscale deep features (MDF)~\cite{LiYu15}.
   }
\label{fig:smaps}
\end{figure*}

% \usepackage{multirow}
\begin{table*}[t]
\centering
\resizebox{1.00\textwidth}{!}
{
\begin{tabular}{|c|c|c|c|c|c|c|c|c|c|c|c|c|}
\hline
Data Set                    & Metric & SF    & GC    & DRFI  & PISA  & BSCA                         & LEGS                         & MC                                    & MDF                                   & FCN                                   & DCL                                  & DCL$^+$                                   \\ \hline
                            & maxF   & 0.700 & 0.719 & 0.845 & 0.837 & 0.830                        & {\color[HTML]{333333} 0.870} & {\color[HTML]{32CB00} \textbf{0.894}} & 0.885                                 & 0.864                                 & {\color[HTML]{3166FF} \textbf{0.905}} & {\color[HTML]{FE0000} \textbf{0.916}} \\ \cline{2-13}
\multirow{-2}{*}{MSRA-B}    & MAE    & 0.166 & 0.159 & 0.112 & 0.102 & 0.130                        & {\color[HTML]{333333} 0.081} & {\color[HTML]{32CB00} \textbf{0.054}} & 0.066                                 & 0.096                                 & {\color[HTML]{3166FF} \textbf{0.052}} & {\color[HTML]{FE0000} \textbf{0.047}} \\ \hline
                            & maxF   & 0.590 & 0.588 & 0.776 & 0.753 & 0.723                        & {\color[HTML]{333333} 0.770} & 0.798                                 & 0.861                                 & {\color[HTML]{32CB00} \textbf{0.867}} & {\color[HTML]{3166FF} \textbf{0.892}} & {\color[HTML]{FE0000} \textbf{0.904}} \\ \cline{2-13}
\multirow{-2}{*}{HKU-IS}    & MAE    & 0.173 & 0.211 & 0.167 & 0.127 & 0.174                        & {\color[HTML]{333333} 0.118} & 0.102                                 & {\color[HTML]{32CB00} \textbf{0.076}} & 0.087                                 & {\color[HTML]{3166FF} \textbf{0.054}} & {\color[HTML]{FE0000} \textbf{0.049}} \\ \hline
                            & maxF   & 0.495 & 0.495 & 0.664 & 0.630 & 0.617                        & {\color[HTML]{333333} 0.669} & {\color[HTML]{32CB00} \textbf{0.703}} & 0.694                                 & 0.681                                 & {\color[HTML]{3166FF} \textbf{0.733}} & {\color[HTML]{FE0000} \textbf{0.757}} \\ \cline{2-13}
\multirow{-2}{*}{DUT-OMRON} & MAE    & 0.147 & 0.218 & 0.150 & 0.141 & 0.191                        & {\color[HTML]{333333} 0.133} & {\color[HTML]{32CB00} \textbf{0.088}} & 0.092                                 & 0.131                                 & {\color[HTML]{3166FF} \textbf{0.084}} & {\color[HTML]{FE0000} \textbf{0.080}} \\ \hline
                            & maxF   & 0.493 & 0.539 & 0.690 & 0.660 & {\color[HTML]{333333} 0.666} & {\color[HTML]{333333} 0.752} & 0.740                                 & 0.764                                 & {\color[HTML]{32CB00} \textbf{0.793}} & {\color[HTML]{3166FF} \textbf{0.815}} & {\color[HTML]{FE0000} \textbf{0.822}} \\ \cline{2-13}
\multirow{-2}{*}{PASCAL-S}  & MAE    & 0.240 & 0.266 & 0.210 & 0.196 & 0.224                        & {\color[HTML]{333333} 0.157} & 0.145                                 & 0.145                                 & {\color[HTML]{32CB00} \textbf{0.128}} & {\color[HTML]{3166FF} \textbf{0.113}} & {\color[HTML]{FE0000} \textbf{0.108}} \\ \hline
                            & maxF   & 0.516 & 0.526 & 0.699 & 0.660 & {\color[HTML]{333333} 0.654} & {\color[HTML]{333333} 0.732} & 0.727                                 & 0.785                                 & {\color[HTML]{32CB00} \textbf{0.795}} & {\color[HTML]{3166FF} \textbf{0.829}} & {\color[HTML]{FE0000} \textbf{0.832}} \\ \cline{2-13}
\multirow{-2}{*}{SOD}       & MAE    & 0.267 & 0.284 & 0.223 & 0.223 & 0.251                        & {\color[HTML]{333333} 0.195} & 0.179                                 & {\color[HTML]{32CB00} \textbf{0.155}} & 0.158                                 & {\color[HTML]{3166FF} \textbf{0.129}} & {\color[HTML]{FE0000} \textbf{0.126}} \\ \hline
\end{tabular}
}
\caption{Comparison of quantitative results including maximum F-measure (larger is better) and MAE (smaller is better). The best three results are shown in \color[HTML]{FE0000}\textbf{red}, \color[HTML]{3166FF}\textbf{blue}\color{black}, and \color[HTML]{32CB00}\textbf{green} \color{black}, respectively. %Note that MC~\cite{zhao2015saliency} and LEGS~\cite{wang2015deep} are overrated on the MSRA-B dataset and LEGS~\cite{wang2015deep} is overrated on the PASCAL-S dataset.
}
\label{tab:comp_quantity}
\end{table*}

\begin{figure*}[t]
    \centerline{
    \includegraphics[width = 0.327\textwidth]{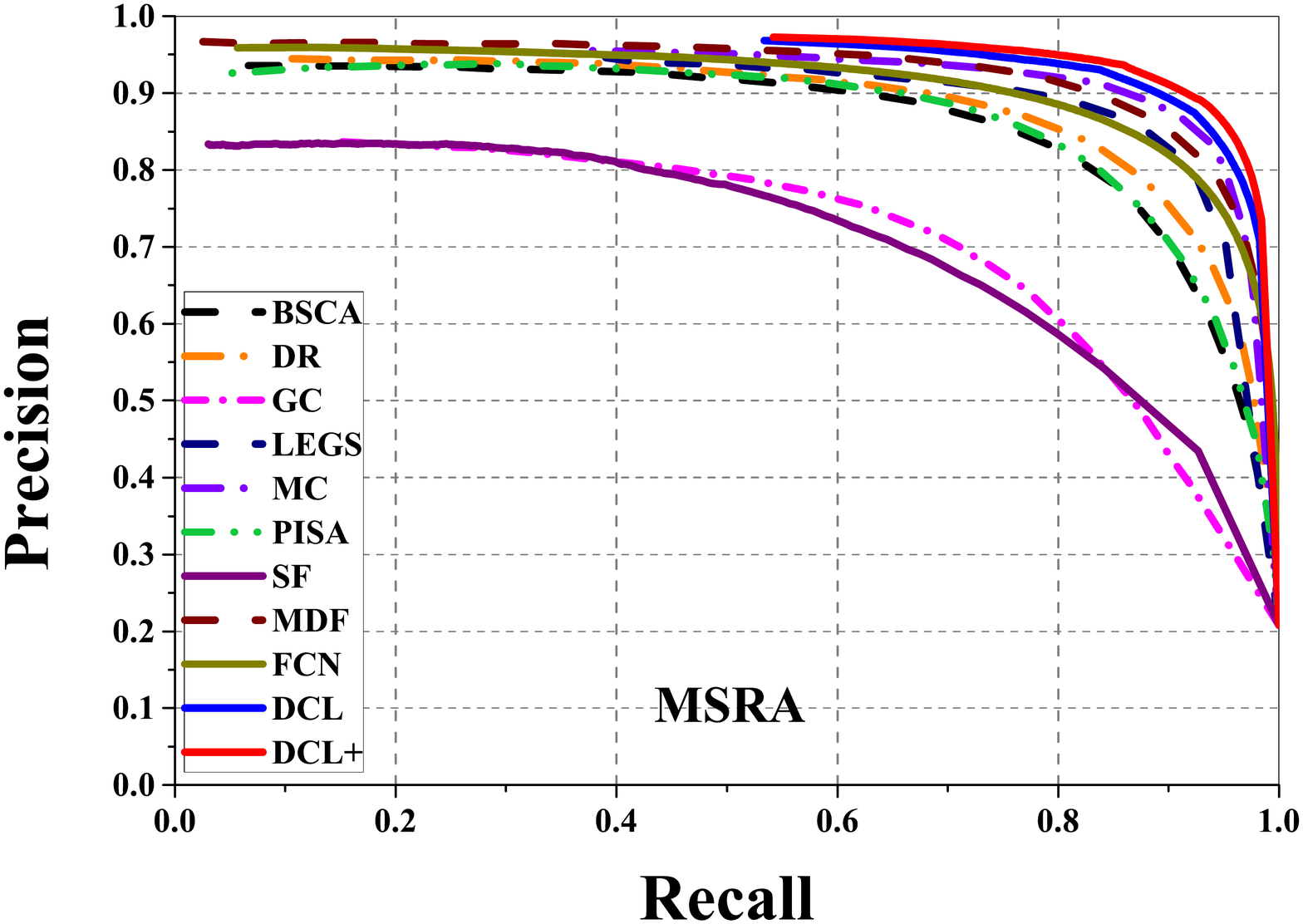}\hfill
    \includegraphics[width = 0.327\textwidth]{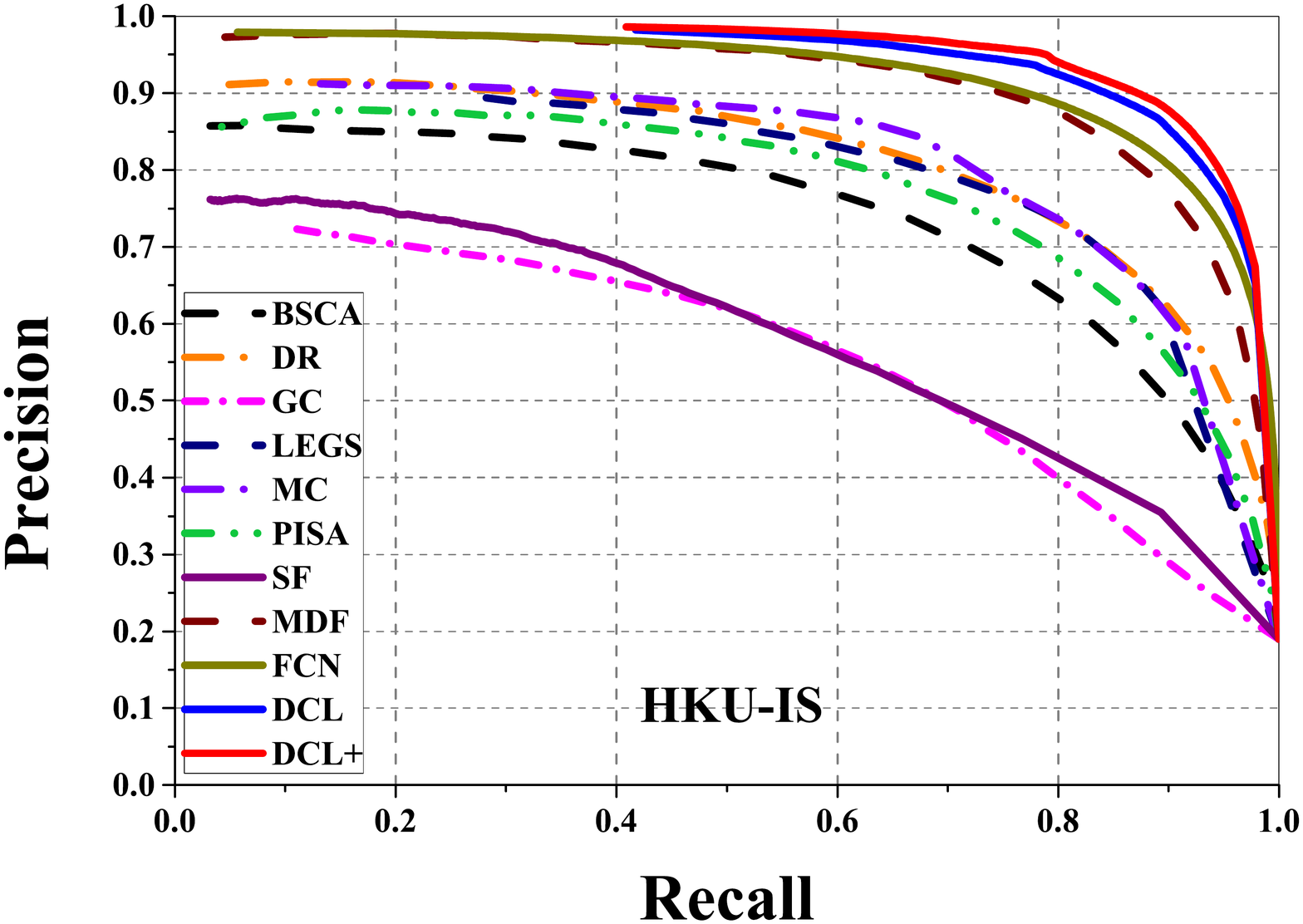}\hfill
    \includegraphics[width = 0.327\textwidth]{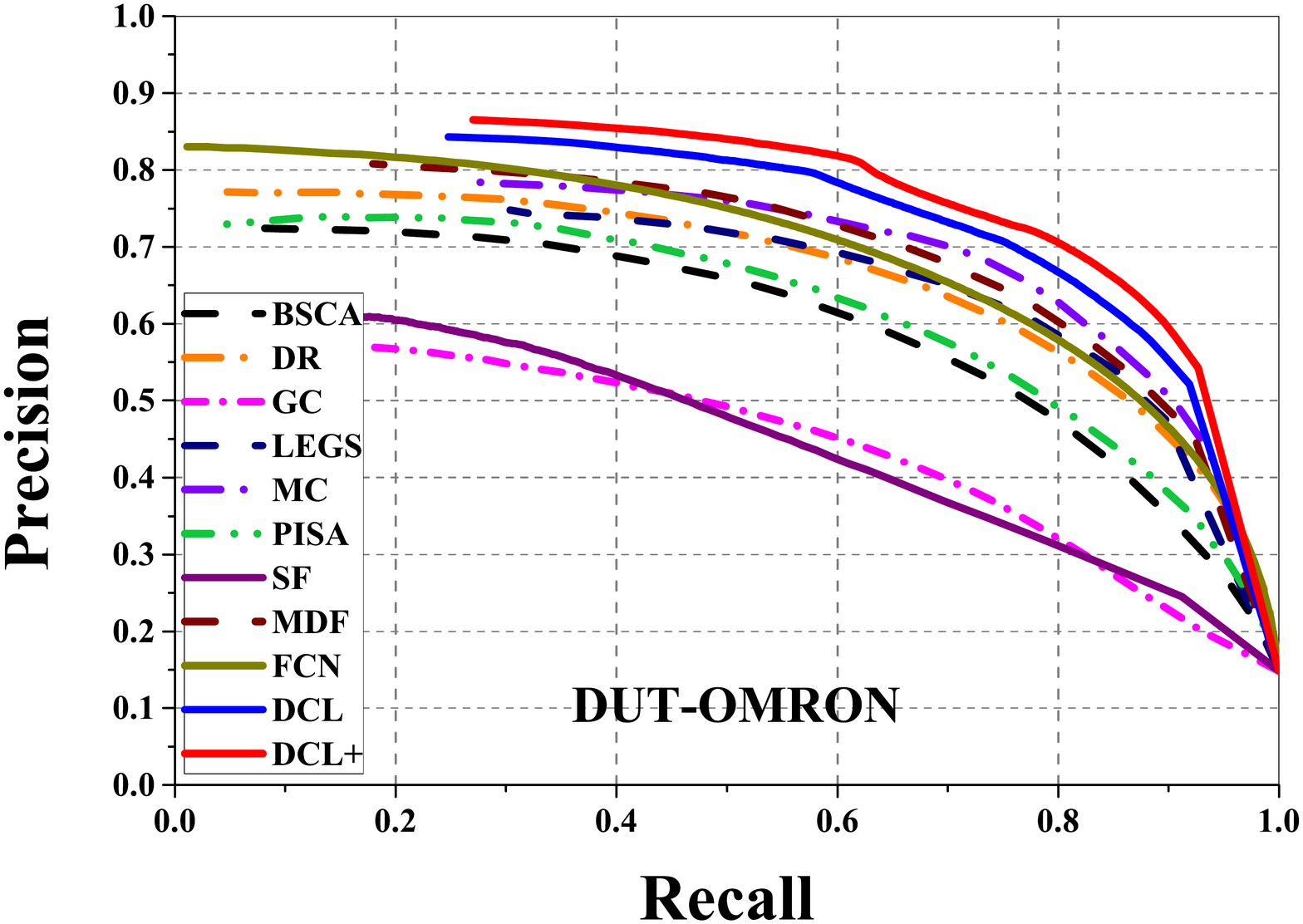}
  }
    %\centerline{
    %\includegraphics[width = 0.333\textwidth]{graphs/dut-omron_pr.eps}\hfill
    %\includegraphics[width = 0.333\textwidth]{graphs/pascal-s_pr.eps}\hfill
    %\includegraphics[width = 0.333\textwidth]{graphs/sod_pr.eps}
  %}
  \caption{Comparison of precision-recall curves of 11 saliency detection methods on 3 datasets. Our DCL and DCL$^+$~(DCL with CRF) consistently outperform other methods across all the testing datasets. Note that MC~\cite{zhao2015saliency} and LEGS~\cite{wang2015deep} are overrated on the MSRA-B dataset and LEGS~\cite{wang2015deep} is also overrated on the PASCAL-S dataset.
  }
  \label{fig:comps_pr}
\end{figure*}
%FCN is a CNN framework originaly proposed for semantic segmentation, we train it

\begin{figure*}[t]
    \centerline{
    \includegraphics[width = 0.327\textwidth]{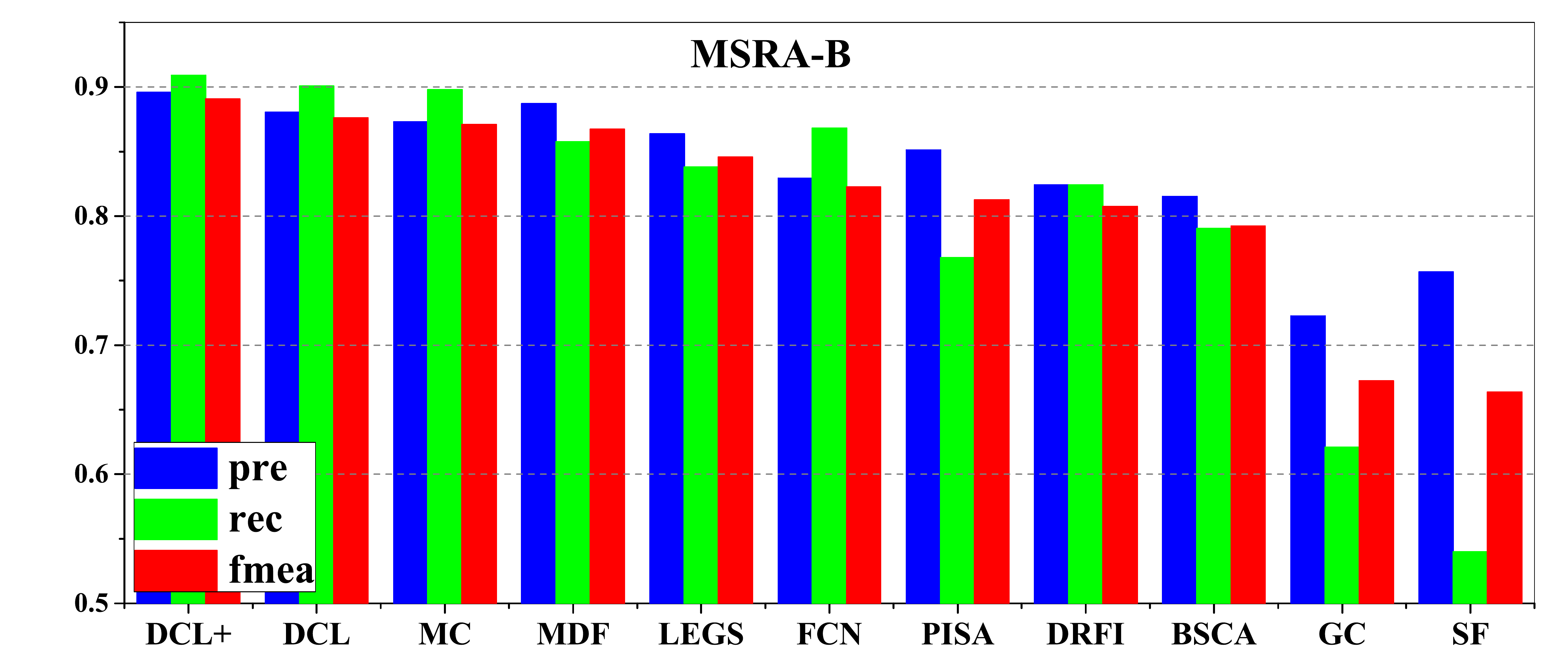}\hfill
    \includegraphics[width = 0.327\textwidth]{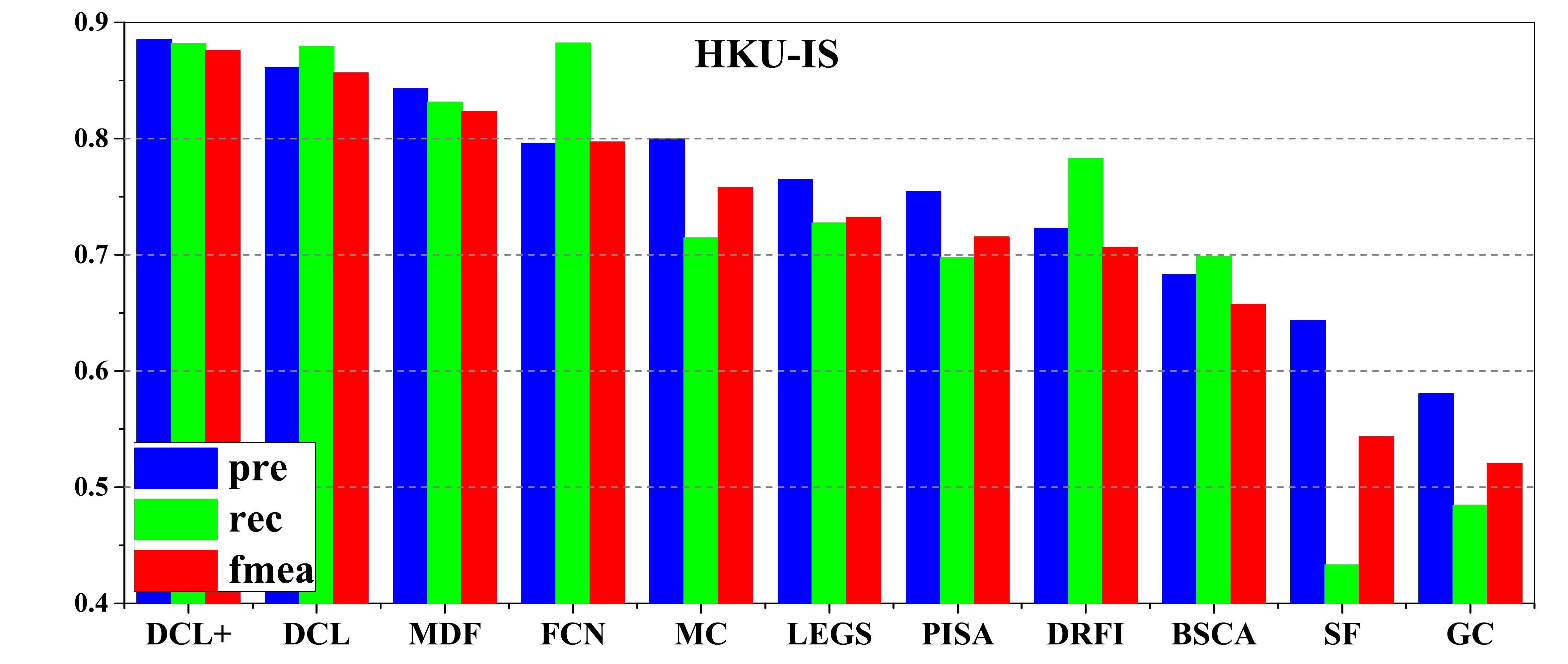}\hfill
    \includegraphics[width = 0.327\textwidth]{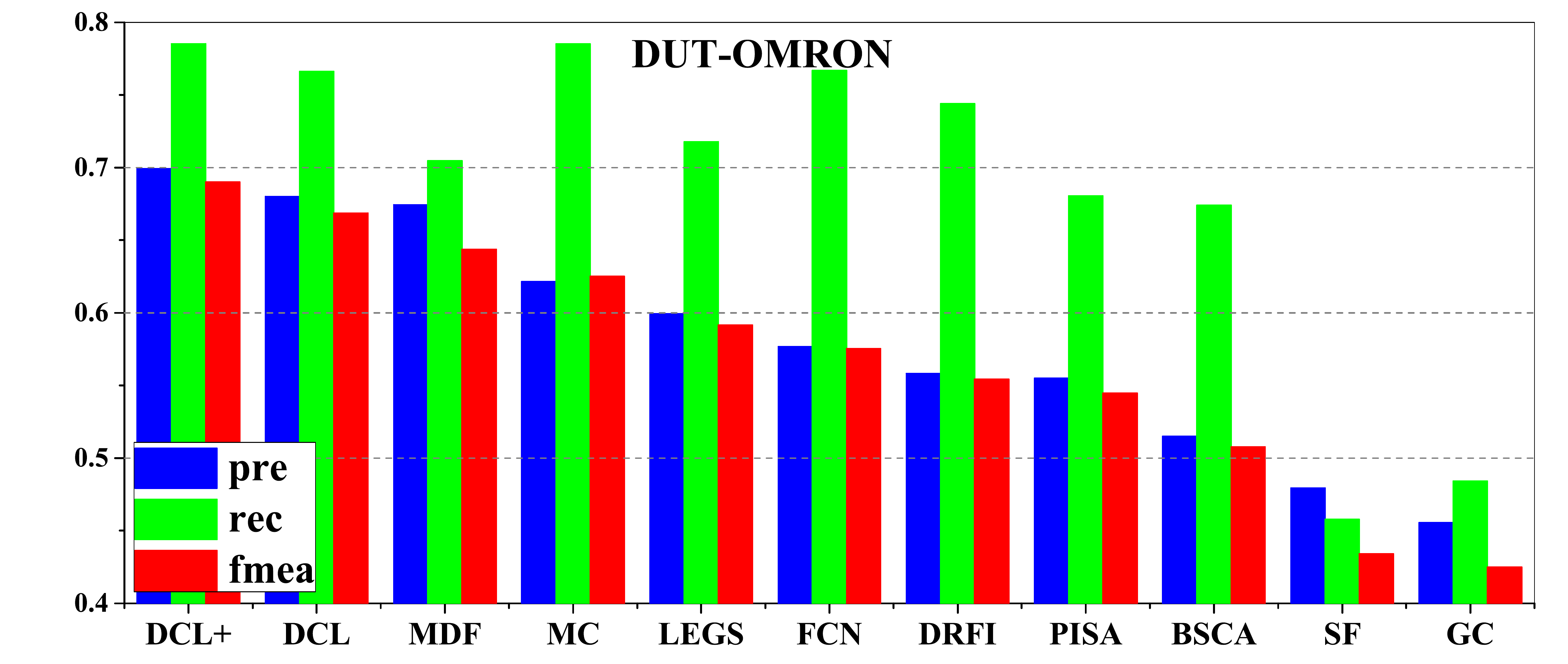}
  }
    %\centerline{
    %\includegraphics[width = 0.333\textwidth]{graphs/dut-omron_prf.eps}\hfill
    %\includegraphics[width = 0.333\textwidth]{graphs/pascal-s_prf.eps}\hfill
    %\includegraphics[width = 0.333\textwidth]{graphs/sod_prf.eps}
  %}
  \caption{Comparison of precision, recall and F-measure (computed using a per-image adaptive threshold) among 11 different methods on 3 datasets. %Our proposed MDF and HDCF consistently occupies the top two F-measure score across all the testing datasets.
  %Note that MC~\cite{zhao2015saliency} and LEGS~\cite{wang2015deep} are overrated on the MSRA-B dataset, and LEGS~\cite{wang2015deep} is also overrated on the PASCAL-S dataset.
  }\label{fig:comps_prf}
\end{figure*}

%\begin{table}[]
%\centering
%\caption{My caption}
%\label{my-label}
%\resizebox{0.48\textwidth}{!}
%{
%\begin{tabular}{llllll}
%Measure     & DCL      & DCL without CRF & MSFCN    & Segment-Level & SC\_MSFCN \\
%MaxFMeasure & 0.917567 & 0.907214        & 0.885579 & 0.874901      & 0.878224  \\
%MAE         & 0.046722 & 0.0520189       & 0.063441 & 0.0571761     & 0.0584488
%\end{tabular}
%}
%\end{table}

\section{Experimental Results}
\subsection{Experimental Setup}
\subsubsection{Datasets}
We evaluate the performance of our method on five public datasets: MSRA-B~\cite{liu2011learning}, %ECSSD~\cite{yan2013hierarchical},
PASCAL-S~\cite{li2014secrets}, DUT-OMRON~\cite{yang2013saliency}, HKU-IS~\cite{LiYu15} and SOD~\cite{martin2001database}. The MSRA-B dataset contains 5,000 images with a variety of image contents. Most of the images has a single salient object. %The ECSSD dataset contains 1,000 structurally complex images from the Internet.
PASCAL-S was built using the validation set of the PASCAL VOC 2010 segmentation challenge. It contains 850 images with the ground truth labeled by 12 subjects. We threshold the masks at 0.5 to obtain binary masks as suggested in~\cite{li2014secrets}. Dut-OMRON contains 5,168 challenging images, each of which has one or more salient objects and relatively complex backgrounds. We have noticed that many saliency annotations in this dataset may be controversial among different human observers. As a result, none of the existing saliency models has achieved a high accuracy on this dataset. HKU-IS is another large dataset containing 4447 challenging images, most of which have either low contrast or multiple salient objects. The SOD dataset contains 300 images and it was originally designed for image segmentation. Many images in this dataset have multiple salient objects with low contrast. All the datasets contain manually annotated groundtruth saliency maps. To facilitate a fair comparison against other methods, we divide the MSRA-B dataset into three parts as in~\cite{jiang2013salient,LiYu15}, 2500 for training, 500 for validation and the remaining 2000 images for testing. To test the adaptability of trained saliency models to other different datasets, we use the models trained on the MSRA-B dataset and test them over all other datasets.

\subsubsection{Evaluation Criteria}
We evaluate the performance using precision-recall (PR) curves, F-measure and mean absolute error (MAE). The precision and recall of a saliency map is computed by converting a continuous saliency map to a binary mask using a threshold and comparing the binary mask against the ground truth. The PR curve of a dataset is obtained from the average precision and recall over saliency maps of all images in the dataset. The F-measure is defined as
\begin{equation}
  F_{\beta} = \frac{(1+\beta^2)\cdot Precision \cdot Recall}{\beta^2\cdot Precision + Recall},
\end{equation}
where $\beta^2$ is set to 0.3 to weigh precision more than recall as suggested in \cite{achanta2009frequency}.  We report the maximum F-measure (maxF) computed from the PR curve. We also report the average precision, recall and F-measure using an adaptive threshold for generating a binary saliency map. The adaptive threshold is determined to be twice the mean value of a saliency map.
%:
%\begin{equation}
% T_a = \frac{2}{W \times H} \sum_{x=1}^{W}\sum_{y=1}^{H}S(x,y),
%\end{equation}
%where $W$ and $H$ are the width and height of the saliency map $S$, and $S(x,y)$ is the saliency score of the pixel at $(x,y)$.
%For binarization, we set all pixels with saliency value larger than $T_a$ as salient and the rest as unsalient.
%We report the average precision, recall and F-measure over each dataset using this adaptive threshold.
In addition, MAE~\cite{perazzi2012saliency} represents the average absolute per-pixel difference between an estimated saliency map and its corresponding ground truth.
%\begin{equation}
%MAE = \frac{1}{W\times H}\sum_{x=1}^{W}\sum_{y=1}^{H}|S(x,y) - G(x,y)|.
%\end{equation}
MAE is meaningful in evaluating the applicability of a saliency model in a task such as object segmentation.

\subsubsection{Implementation}
Our proposed deep contrast network has been implemented on the basis of Caffe~\cite{jia2014caffe}, an open source framework for CNN training and testing.
%More specifically, for the MS-FCN, we take the VGG16 model pre-trained on ImageNet, and fine-tune it on a saliency dataset using stochastic gradient descent and the cross-entropy loss function described in Section~\ref{sec:dcn_train}.
We resize all the images %and groundtruth saliency maps
to $321\times 321$ pixels for training, and set the initial learning rate to 0.01 for all newly added layers with one channel and 0.001 for all other layers. The momentum parameter is set to 0.9 and the weight decay is 0.0005. For the segment-level stream, the number of superpixels is set to 400 with 3 different scales~(200, 150 and 50 respectively). A $2\times 2$ grid is used for spatial pooling over each segment. Thus the aggregated feature for each segment has 6144 dimensions, and this feature is further fed into two fully connected layers each of which has 300 neurons. %We refer to~\cite{LiYu15} for the learning parameters setting.
The parameters of the fully connected CRF are determined through cross validation as in~\cite{krahenbuhl2012efficient} on the validation set and finally the paramters of $w_1$, $w_2$, $\sigma_\alpha$, $\sigma_\beta$, and $\sigma_\gamma$ are set to $3.0$, $5.0$, $3.0$, $50.0$ and $3.0$ respectively in our experiments.

We use DCL to denote our saliency model based on deep contrast learning only without CRF-based post-processing, and DCL$^+$ to denote the saliency model that includes CRF-based refinement.
While it takes around 25 hours to train our deep contrast network using the MSRA-B dataset, it only takes 1.5 seconds for the trained model (DCL) to detect salient objects in a testing image with 400x300 pixels on a PC with an NVIDIA Titan Black GPU and a
3.4GHz Intel processor. Note that this is far more efficient than the latest deep learning based methods which treat all image patches as independent data samples for saliency regression. % and bring lots of computation redundancy.
%However,
CRF-based post-processing %is more expensive, and
requires additional 0.8 second per image. %since we need to compute generalized eigenvectors used in the CRF model.
Experimental results will show that DCL alone without CRF-based post-processing already outperforms existing state-of-the-art methods.

\subsection{Comparison with the State of the Art}
We compare our saliency models (DCL and DCL$^+$) against eight recent state-of-the-art methods, including SF~\cite{perazzi2012saliency}, GC~\cite{cheng2015global}, %HS~\cite{yan2013hierarchical},
DRFI~\cite{jiang2013salient}, PISA~\cite{pisa15PAMI}, BSCA~\cite{qin2015saliency}, LEGS~\cite{wang2015deep}, MC~\cite{zhao2015saliency} and MDF~\cite{LiYu15}. The last three are the latest deep learning based methods. For fair comparison, we use either the implementations or the saliency maps provided by the authors. In addition, we also train a fully convolutional neural network (FCN) (the FCN-8s network proposed in ~\cite{long2014fully}) for comparison. To train the FCN saliency model, we simply replace its last softmax layer with a sigmoid cross-entropy layer for saliency inference, and fine-tune the revised model using the training sets in the aforementioned saliency datasets.
%We compare our proposed saliency model~(DCL) with nine latest state-of-the-art methods, including saliency filters~(SF)~\cite{perazzi2012saliency}, global contrast~(GC)~\cite{cheng2015global}, hierarchical saliency~(HS)~\cite{yan2013hierarchical}, discriminative regional feature integration (DRFI)~\cite{jiang2013salient}, pixelwise image saliency aggregating~(PISA)~\cite{pisa15PAMI}, single-layer cellular automata~(BSCA)~\cite{qin2015saliency}, local estimation and global search based deep networks~(LEGS)~\cite{wang2015deep}, multi-context deep learning~(MC)~\cite{zhao2015saliency}, and multiscale deep features (MDF)~\cite{LiYu15}. The latter three methods are deep learning based methods. We also train a fully convolutional neural network~(FCN) based on the FCN-8s network proposed in ~\cite{long2014fully} for comparison. To train the FCN saliency model, we just modify the last softmax layer to a sigmoid cross-entropy layer for saliency inference, and fine-tune the model using the same training dataset as we train our model. For fair comparison, we use either the implementations or the saliency maps provided by the authors.

A visual comparison is shown in Figure~\ref{fig:smaps}. As can be seen, our method generates more accurate saliency maps in various challenging cases, e.g., objects touching the image boundary (the first two rows), multiple disconnected salient objects (the middle two rows) and low contrast between object and background (the last two rows). It is necessary to point out that the performance of MC~\cite{zhao2015saliency} is overrated on the MSRA-B dataset and the performance of LEGS~\cite{wang2015deep} is overrated on both the MSRA-B dataset and the PASCAL-S dataset because most testing images in the corresponding datasets were used as training samples for the publicly released trained models of MC and LEGS used in our comparison.

Our method significantly outperforms all existing salient object detection algorithms across the aforementioned public datasets in terms of PR curve~(Fig.~\ref{fig:comps_pr}) and average precision, recall and F-measure (Fig.~\ref{fig:comps_prf}). Refer to the supplemental materials for the results on the PASCAL-S and SOD datasets. Moreover, we report a quantitative comparison w.r.t. maximum F-measure and MAE in Table~\ref{tab:comp_quantity}. Our complete model~(DCL$^+$) improves the maximum F-measure achieved by the best-performing existing algorithm by 3.5\%, %7.7\%,
5.0\%, 7.7\%, 7.6\% and 6.0\% respectively on MSRA-B (skipping MC and LEGS on this dataset), %ECSSD,
HKU-IS, DUT-OMRON, PASCAL-S (skipping LEGS on this dataset) and SOD. And at the same time, our model lowers the MAE by 28.8\%, %32.2\%,
35.5\%, 9.1\%, 25.5\% and 18.7\% respectively on MSRA-B (skipping MC and LEGS on this dataset), %ECSSD,
HKU-IS, DUT-OMRON, PASCAL-S (skipping LEGS on this dataset) and SOD. We can also see that our model without CRF (DCL) significantly outperforms all evaluated salient object detection algorithms across all the considered datasets. Our model also significantly outperforms the FCN adapted from a model originally designed for semantic segmentation~\cite{long2014fully} because we explicitly perform deep contrast learning, which is critical for saliency detection.

\begin{figure}[ht]
    \centerline{
    \includegraphics[width = 0.24\textwidth]{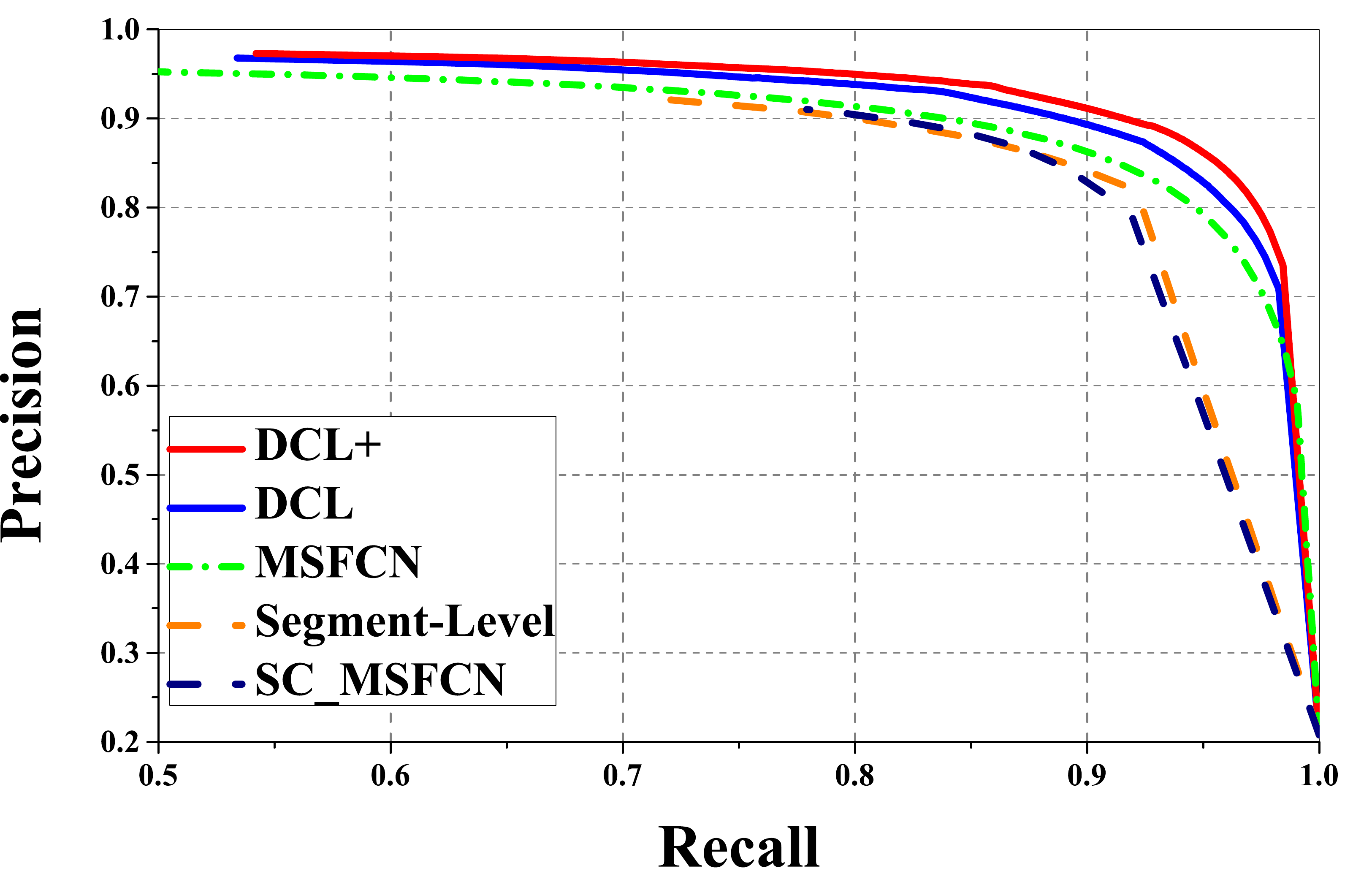}\hfill
    \includegraphics[width = 0.24\textwidth]{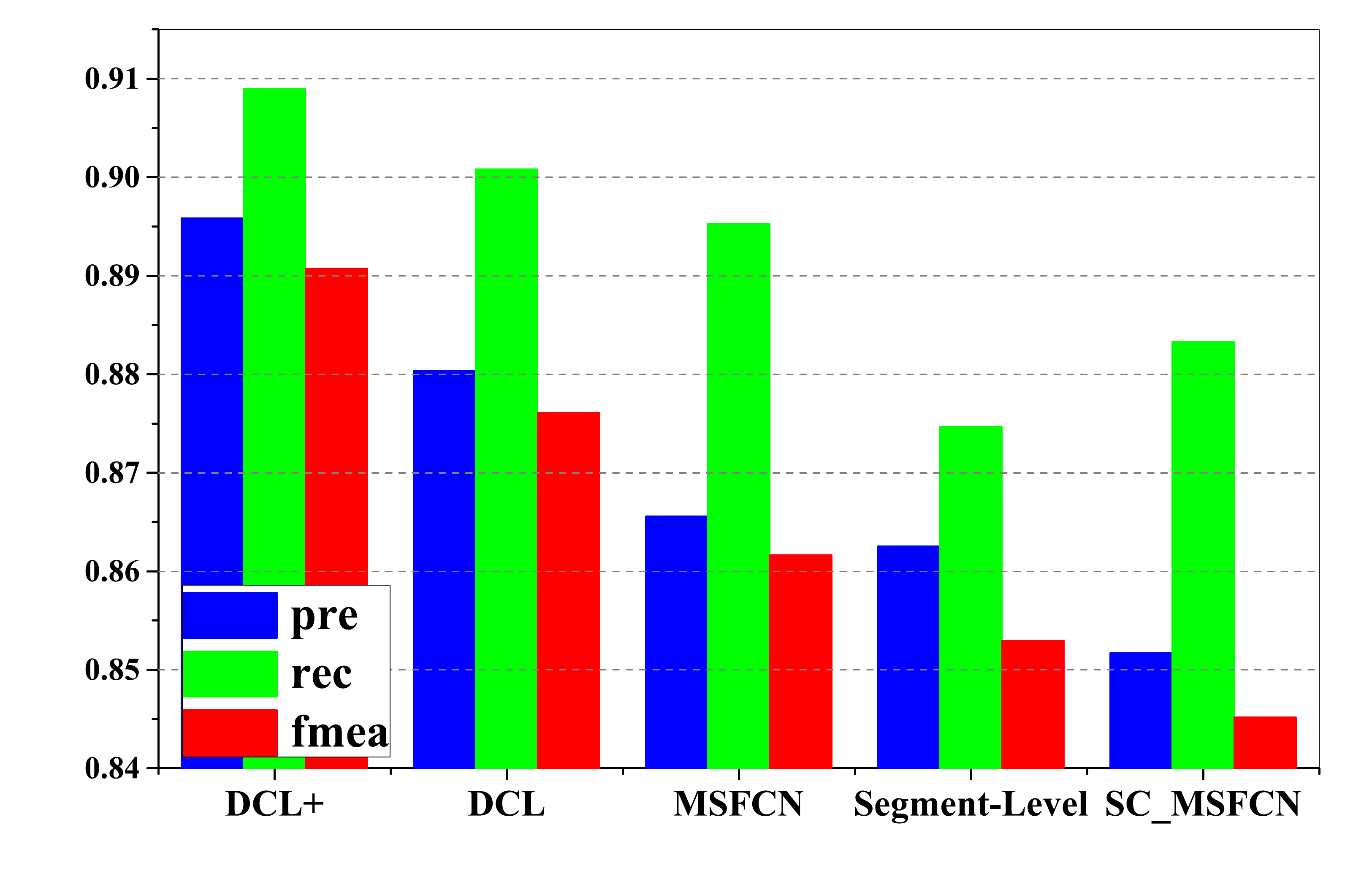}\hfill
  }\vspace{-0mm}
    %\centerline{\hfill (a) \hfill\hfill (b) \hfill\hfill (c) \hfill\hfill (d) \hfill}\vspace{-1mm}
    \caption{Componentwise efficacy of the proposed deep contrast network and the effectiveness of the CRF model.}
\label{fig:ablation_study}
\end{figure}

\subsection{Ablation Studies}
\subsubsection{Effectiveness of Deep Contrast Network}
Our deep contrast network consists of a fully convolutional stream and a segment-wise spatial pooling stream. To show the effectiveness and necessity of these two components, we compare the saliency map $S_1$ generated from the first stream (MS-FCN), the saliency map $S_2$ from the second segment-level stream and the fused saliency map from $S_1$ and $S_2$ (DCL) using testing images in the MSRA-B dataset. As shown in Fig.~\ref{fig:ablation_study}, the fused saliency map (DCL) consistently achieves the best performance on average precision, recall and F-measure, and the fully convolutional stream (MS-FCN) has more contribution to the fused result than the segment-wise spatial pooling stream. These two streams are complementary to each other, and our trained deep contrast network is capable of discovering and understanding subtle visual contrast among multi-scale feature maps as well as between neighboring segments. To demonstrate the effectiveness of MS-FCN, we also generate saliency maps from the last scale of MS-FCN (the best performing scale) for comparison. The last scale of MS-FCN is in fact the fully convolutional version of the original VGG16 network. As shown in Fig.~\ref{fig:ablation_study}, this single scale of MS-FCN (called SC\_MSFCN) performs much worse than the complete version of MS-FCN in terms of the PR curve as well as the average precision, recall and F-measure.

\subsubsection{Effectiveness of CRF}\label{sec:effectiveness_spatial_coherence}
In Section~\ref{sec:spatial_coherence}, a fully connected CRF is incorporated to improve the spatial coherence of the saliency maps from our deep contrast network. To validate its effectiveness, we have also evaluated the performance of our final saliency model with and without the CRF using the testing images in the MSRA-B dataset. The results are also shown in Fig.~\ref{fig:ablation_study}. It is evident that the CRF improves the accuracy of our model.

\section{Conclusions}
In this paper, we have introduced an end-to-end deep contrast network for salient object detection. Our deep network consists of two complementary components, a pixel-level fully convolutional stream and a segment-level spatial pooling stream.
%The first stream takes the raw image as input and directly produces a saliency map with pixel-level accuracy. The second stream extracts segment-wise features very efficiently, and better models saliency discontinuities along object boundaries. Finally,
A fully connected CRF model can be optionally incorporated to further improve spatial coherence and contour localization in the fused result from these two streams. Experimental results demonstrate that our deep model can significantly improve the state of the art.

\section*{Acknowledgment}
The first author is supported by Hong Kong Postgraduate Fellowship.

{%\small
\bibliographystyle{ieee}
\bibliography{n2n_saliency_arxiv}
}

\end{document}